\documentclass{article}

\usepackage{microtype}
\usepackage{graphicx}
\usepackage{subcaption}
\usepackage{booktabs} % for professional tables

\usepackage{hyperref}

\usepackage[accepted]{icml2024}

\usepackage{amsmath}
\usepackage{amssymb}
\usepackage{mathtools}
\usepackage{amsthm}
\usepackage{multirow}
\usepackage{xspace}

\usepackage[capitalize,noabbrev]{cleveref}

\theoremstyle{plain}

\theoremstyle{definition}

\theoremstyle{remark}

\usepackage[disable,textsize=tiny]{todonotes}
\usepackage{arydshln}

\usepackage{tikz}
\usetikzlibrary{decorations.pathreplacing, calligraphy}
\usetikzlibrary{shapes.callouts}
\usetikzlibrary{chains, fit, quotes, automata}
\usetikzlibrary{arrows}
\usetikzlibrary{matrix,positioning}
\usetikzlibrary{shapes.geometric, shapes.misc}
\usetikzlibrary{calc}
\usepackage{tikzpeople}
\usepackage{fontawesome5}
\usepackage{enumitem}

\newcommand{\method}{CatLIP\@\xspace}
\newcommand{\imagenet}{ImageNet-1k\@\xspace}
\newcommand{\datacomp}{DataComp-1.3B\@\xspace}
\newcommand{\places}{Places365\@\xspace}
\newcommand{\cctm}{CC3M\@\xspace}
\newcommand{\coco}{COCO\@\xspace}

\definecolor{deemph}{gray}{0.6}
\newcommand{\gc}[1]{\textcolor{deemph}{#1}}

\icmltitlerunning{\method: CLIP-level Visual Recognition Accuracy with 2.7\texttimes{} Faster Pre-training on Web-scale Image-Text Data}

\begin{document}

\twocolumn[
%\icmltitle{Simplicity Meets Scalability and Efficiency: Multi-label Image Pre-training with Weak Image-Text Pairs}
\icmltitle{\method: CLIP-level Visual Recognition Accuracy with 2.7\texttimes{} Faster Pre-training on Web-scale Image-Text Data}

%\icmltitle{\method:  2.7\texttimes{} Faster Pre-training  with CLIP-level  Visual Recognition Accuracy on Web-scale Image-Text Data}

% It is OKAY to include author information, even for blind
% submissions: the style file will automatically remove it for you
% unless you've provided the [accepted] option to the icml2024
% package.

% List of affiliations: The first argument should be a (short)
% identifier you will use later to specify author affiliations
% Academic affiliations should list Department, University, City, Region, Country
% Industry affiliations should list Company, City, Region, Country

% You can specify symbols, otherwise they are numbered in order.
% Ideally, you should not use this facility. Affiliations will be numbered
% in order of appearance and this is the preferred way.
\icmlsetsymbol{equal}{*}

\begin{icmlauthorlist}
\icmlauthor{Sachin Mehta}{apple}
\icmlauthor{Maxwell Horton}{apple}
\icmlauthor{Fartash Faghri}{apple}
\icmlauthor{Mohammad Hossein Sekhavat}{apple}
\icmlauthor{Mahyar Najibi}{apple}
\icmlauthor{Mehrdad Farajtabar}{apple}
\icmlauthor{Oncel Tuzel}{apple}
%\icmlauthor{}{sch}
\icmlauthor{Mohammad Rastegari}{apple}
\end{icmlauthorlist}

\icmlaffiliation{apple}{Apple}

% You may provide any keywords that you
% find helpful for describing your paper; these are used to populate
% the "keywords" metadata in the PDF but will not be shown in the document
\icmlkeywords{Machine Learning, ICML}

\vskip 0.3in
]

% this must go after the closing bracket ] following \twocolumn[ ...

% This command actually creates the footnote in the first column
% listing the affiliations and the copyright notice.
% The command takes one argument, which is text to display at the start of the footnote.
% The \icmlEqualContribution command is standard text for equal contribution.
% Remove it (just {}) if you do not need this facility.

\printAffiliationsAndNotice{Sachin Mehta led this project and made significant technical contributions.}  % leave blank if no need to mention equal contribution
%\printAffiliationsAndNotice{\icmlEqualContribution} % otherwise use the standard text.

\begin{abstract}
Contrastive learning has emerged as a transformative method for learning effective visual representations through the alignment of image and text embeddings. However, pairwise similarity computation in contrastive loss between image and text pairs poses computational challenges.  This paper presents a novel weakly supervised pre-training of vision models on web-scale image-text data. The proposed method reframes pre-training on image-text data as a classification task. Consequently, it eliminates the need for pairwise similarity computations in contrastive loss, achieving a remarkable $2.7\times$ acceleration in training speed compared to contrastive learning on web-scale data. Through extensive experiments spanning diverse vision tasks, including detection and segmentation, we demonstrate that the proposed method maintains high representation quality. Our source code along with pre-trained model weights and training recipes is available at \url{https://github.com/apple/corenet}.
\end{abstract} 

\section{Introduction}
\label{sec:intro}

\begin{figure}[t!]
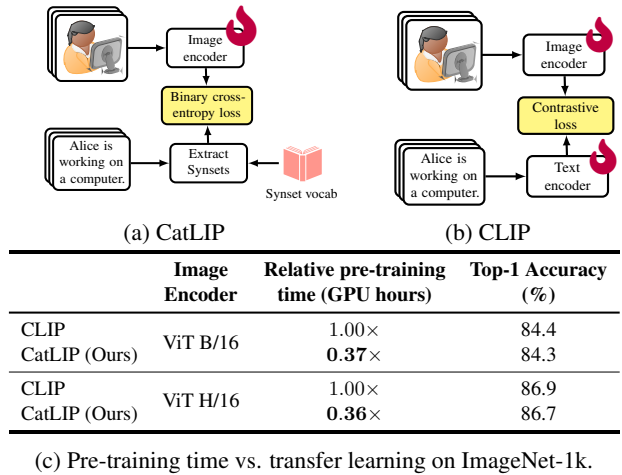

    \centering
    \begin{subfigure}[b]{0.55\columnwidth}
        \centering
         \resizebox{\columnwidth}{!}{
             \input{tikz_figures/overview.tikz}\catLip
         }
         \caption{\method}
         \label{fig:overview_catlip}
     \end{subfigure}
     \hfill
     \begin{subfigure}[b]{0.44\columnwidth}
        \centering
         \resizebox{\columnwidth}{!}{
             \input{tikz_figures/overview.tikz}\clip
         }
         \caption{CLIP}
         \label{fig:overview_clip}
     \end{subfigure}
     \vfill
     \begin{subfigure}[b]{\columnwidth}
        \resizebox{\columnwidth}{!}{
        \begin{tabular}{lccc}
            \toprule[1.5pt]
             & \textbf{Image} & \textbf{Relative pre-training} & \textbf{Top-1 Accuracy} \\
             & \textbf{Encoder} & \textbf{time (GPU hours)} & \textbf{(\%)} \\
            \midrule[1pt]   
             CLIP & \multirow{2}{*}{ViT B/16} & $1.00 \times$ & 84.4 \\
             \method (Ours) &  & $\mathbf{0.37 \times}$ & 84.3 \\
            \midrule
            CLIP & \multirow{2}{*}{ViT H/16} & $1.00\times$ & 86.9 \\
             \method (Ours) & & $\mathbf{0.36 \times}$ & 86.7 \\
             \bottomrule[1.5pt]
        \end{tabular}
        }
        \caption{Pre-training time vs. transfer learning on \imagenet.}
        \label{fig:catlip_clip_compare_main_results}
     \end{subfigure}
    \caption{\textbf{\method is $\mathbf{2.7\times}$ faster to pre-train than CLIP while maintaining down-stream accuracy.} For a fair comparison, we calculate GPU hours by training \method and CLIP with a batch size of 65k for one epoch of \datacomp on the same hardware configuration. Finetuning accuracy of \method and CLIP is reported on \imagenet dataset. Here, \textcolor{purple}{\faFire} represents trainable backbones.}
    \label{fig:clip_catlip_main}
\end{figure}

\begin{figure*}[t!]
    \centering
    \begin{subfigure}[b]{0.66\columnwidth}
        \includegraphics[height=95px]{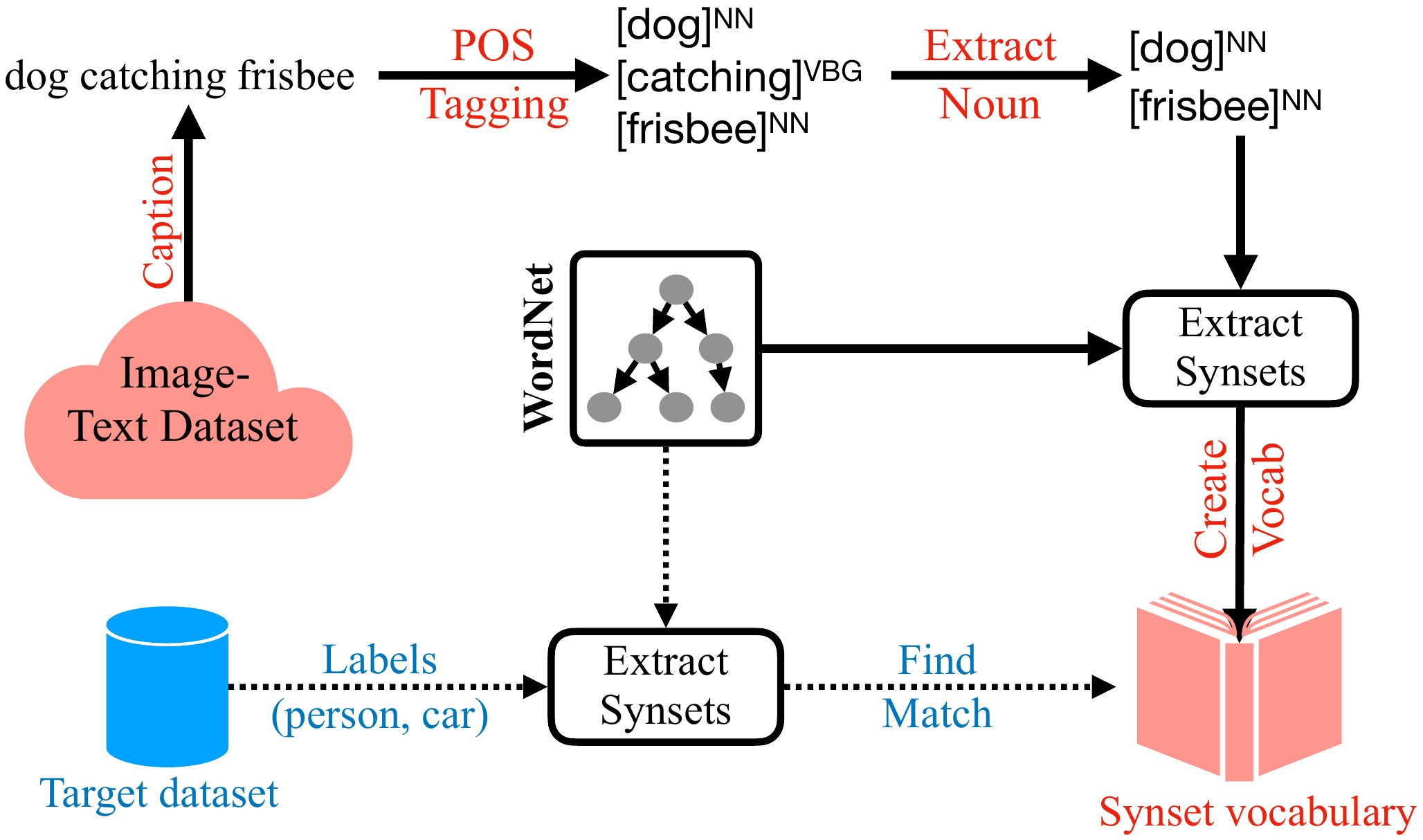}
        \caption{}
        \label{fig:zero_shot_ann_app_main}
    \end{subfigure}
    \hfill
    \begin{subfigure}[b]{0.66\columnwidth}
         \centering
         \includegraphics[height=95px]{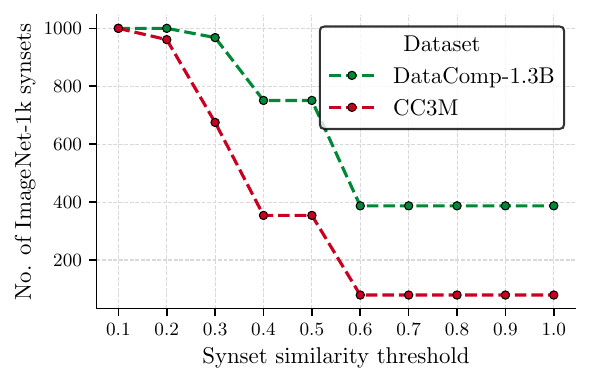}
         \caption{}
         \label{fig:tau_500_main}
     \end{subfigure}
      \hfill
    \begin{subfigure}[b]{0.66\columnwidth}
         \centering
         \includegraphics[height=95px]{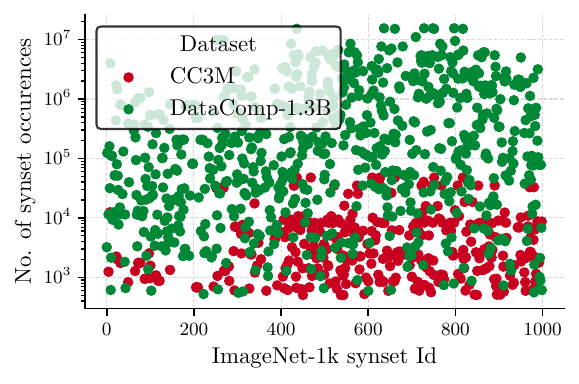}
         \caption{}
         \label{fig:in1k_syn_freq_pretraining_data_main}
    \end{subfigure}
    \caption{\imagenet labels appear frequently in image-text datasets, contributing to increased zero-shot classification accuracies of CLIP models. (a) shows the process of finding labels of the target dataset in the image-text dataset. (b) shows the number of \imagenet synsets at different similarity thresholds between synset vocabulary and \imagenet synset. Exact match (similarity score of 1.0) for approximately 40\% of \imagenet labels is found in the \datacomp captions. (c) illustrates occurrences of \imagenet synsets in image-text datasets, with larger datasets exhibiting more samples per synset. Here, the vocabulary pruning threshold, $V_\tau$, is set to $500$.}
    \label{fig:main_figure}
\end{figure*}

The pre-training of computer vision models on large-scale datasets has enabled models to learn intricate patterns, features, and representations, leading to their effective generalization across diverse visual tasks \citep[e.g.,][]{krizhevsky2012imagenet,kornblith2019better,kirillov2023segment}. The pre-training methods offer different trade-offs.  Fully supervised methods benefit from well-defined optimization objectives, enabling the model to learn strong semantic representations \citep{zhai2022scaling,dehghani2023scaling}. However, obtaining large-scale manually labeled datasets for such methods is time-consuming and expensive. In contrast, unsupervised methods offer scalability to large datasets as they do not rely on manual labels \cite{chen2020simple,he2022masked}. Yet, achieving competitive performance compared to supervised methods often requires the use of either larger models or longer training. Weakly-supervised approaches strike a balance by incorporating noisy supervision (e.g., captions associated with an image) while maintaining scalability \cite{singh2022revisiting, jia2021scaling}. Notably, contrastive learning using weakly labeled data is a transformative method, especially with the increasing availability of both small- and large-scale weakly labeled image-text datasets in the public domain \cite{thomee2016yfcc100m,sharma2018conceptual,gadre2023datacomp,schuhmann2022laion}.

Contrastive Language-Image Pretraining (CLIP)~\citep{radford2021learning} aligns image-text pairs by encouraging the similarity between input image-text pairs and penalizes similarity between unpaired samples. The process involves considering multiple negative pairs for every positive pair, leading to a large number of pairwise comparisons and making it computationally intensive. In comparison, supervised learning approaches leverage simpler and more computationally efficient objective functions, such as the cross-entropy loss. This avoids the need for a global collection of negative samples for effective learning, addressing computational challenges inherent in contrastive learning. \citet{zhai2023sigmoid} reduce the computational overhead of contrastive loss in CLIP by reducing the dependence of the loss on the entire batch. However, they still have to compute pairwise similarities.

Effectively harnessing the advantages of both web-scale image-text datasets and image classification frameworks for faster training while preserving representation quality remains an open question. We begin our investigation by looking into the presence of \imagenet \cite{deng2009imagenet} classes in both small- and large-scale image-text datasets that are commonly used for training CLIP. Many of the \imagenet classes emerge within the \cctm \cite{sharma2018conceptual} and \datacomp \cite{gadre2023datacomp} captions, often appearing as either exact matches or closely related concepts (\cref{fig:tau_500_main}). Moreover, they manifest in multiple instances, underscoring the knowledge of objects within these datasets (\cref{fig:in1k_syn_freq_pretraining_data_main}).

Based on these observations, we introduce a method that addresses the trade-off between efficiency and scalability on a weakly labeled web-scale image-text dataset (\cref{fig:overview_catlip}). In contrast to contrastive pre-training, we conceptualize image-text pre-training as a classification problem where multi-labels are obtained by extracting nouns from text captions. We call the proposed method, \method (\textbf{Cat}egorical \textbf{L}oss for \textbf{I}mage-text \textbf{P}re-training). \cref{fig:catlip_clip_compare_main_results} shows that \method is $2.7\times$ more efficient to train than CLIP while preserving downstream performance on the \imagenet dataset. We present extensive experiments in \cref{sec:scale,sec:generalize} to demonstrate the effectiveness of the proposed method.

The main contributions of our paper are: 
\begin{enumerate}[leftmargin=*]
    \item We introduce a novel approach for accelerating the pre-training of vision models on image-text data (\cref{sec:simple}). To the best of our knowledge, this is the first method to cast pre-training on image-text data as classification task.
    \item \method improves the accuracy with  data and model scaling (\cref{sec:scale}). Noteworthy are the results on small-scale image-text data, where the performance of model improves with longer training when trained with \method as compared to CLIP (\cref{ssec:catlip_pretrianing}).
    \item Standard approach for transfer learning involves initializing the model backbone with pre-trained weights and randomly initializing the classifier. Because target labels can be a subset of \method's vocabulary, it enables the extraction of embeddings associated with target task labels from the classification layer of the pre-trained model. These embeddings can then be utilized to initialize the classification layer in the target task, facilitating data-efficient transfer learning (\cref{ssec:model_scaling}; \cref{fig:data_efficiency_lp}).    
    \item Through extensive experiments spanning various downstream tasks, including object detection and semantic segmentation (refer to Section \ref{sec:generalize}), we demonstrate the effectiveness of representations learned by \method, showing comparable performance to CLIP. As an instance, Mask R-CNN \cite{he2017mask} employing a vision transformer (ViT B/16; \cite{dosovitskiy2020image}) backbone achieved a mean average precision score of 49.9 on COCO \cite{lin2014microsoft} for both \method and CLIP. We highlight that CLIP requires $2.7\times$ more time for pre-training compared to \method on \datacomp (\cref{fig:catlip_clip_compare_main_results}).
\end{enumerate}

\section{Related Work}
\label{sec:related_work}

\paragraph{Learning visual representations at scale.} Several studies have demonstrated the benefits of pre-training models on large-scale datasets. The first line of research focuses on leveraging large-scale labeled datasets for training vision models (e.g., \imagenet, JFT 300M \cite{dosovitskiy2020image}, and JFT-3B \cite{zhai2022scaling}). This approach relies on explicit supervision, enabling models to acquire robust semantic representations. However, it comes with the drawback of requiring extensive manual labeling for constructing large-scale datasets. The second line of research focuses on leveraging unlabeled data \cite{chen2020simple,he2022masked}. While these approaches offer scalability, achieving competitive performance against supervised models often necessitates the use of larger models or longer training. The third line of research focuses on weak supervision \cite{radford2021learning, jia2021scaling, singh2022revisiting}, aiming to strike a trade-off between fully supervised learning, which provides good accuracy, and unsupervised learning, which offers scalability. 

Our work falls in the category of weakly-supervised learning from image-text data. Unlike previous works that align image-text pairs using contrastive learning, we formulate learning on image-text datasets as a classification problem. This paradigm shift in pre-training allows us to address the computational bottlenecks in contrastive learning and reduce pre-training time by $2.7\times$ while preserving accuracy on downstream tasks.

\paragraph{Efficient contrastive image-text pre-training.} The majority of methods for pre-training on image-text data follow contrastive learning \citep[e.g.,][]{radford2021learning, jia2021scaling, xu2021videoclip, schuhmann2022laion,cherti2023reproducible}. However, this approach is computationally expensive due to the computation of pairwise similarities for each image and text embedding. While some efforts have been made to enhance the training efficiency of image-text pre-training, they may not necessarily address the specific bottlenecks associated with contrastive learning.

\emph{Pre-trained encoders.} LiT \cite{zhai2022lit} aligns image and text embeddings with a frozen image encoder and a learnable text encoder. APE \cite{rosenfeld2022ape} uses frozen image and text encoders to extract image and text embeddings, which are then aligned using a small multi-layer perceptron. Although these methods may not explicitly tackle the computational challenges associated with contrastive learning, they reduce alignment time by optimizing fewer parameters, and are complementary to our work.

\emph{ViT-specific optimizations.} FLIP \cite{li2023scaling} extends masked autoencoders \cite{he2022masked} for contrastive learning. It involves randomly masking image tokens and subsequently aligning text tokens with unmasked image tokens using contrastive learning. Hardware-level optimizations \cite{dao2022flashattention, rabe2021self} further reduce the memory footprint of attention in transformers and can aid in accelerating CLIP training. While this model-specific strategy accelerates training, it doesn't directly tackle the bottlenecks associated with contrastive learning. These methods complement our work, offering potential enhancements in training efficiency for specific models.

\emph{Better objective functions.} SigLIP \cite{zhai2023sigmoid} mitigated the computational overhead of the softmax-based contrastive loss in CLIP by calculating local pairwise similarities through a sigmoid loss. However, the extent to which local sigmoid loss enhances training efficiency in comparison to softmax-based loss remains unclear. UniCL \cite{yang2022unified} tries to unify supervised and weakly-supervised datasets by constructing image-text-label triplets.  While it can improve accuracy, it still needs to compute pair-wise similarities, making the pre-training slower and more expensive. 

In contrast to existing contrastive image-text pre-training approaches, our work reframes image-text pre-training as a classification task, significantly reducing pre-training time while preserving CLIP-level accuracy in downstream tasks.

\paragraph{Improving large-scale pre-training.} There are works, not specifically tailored to image-text pre-training, that aim to enhance pre-training efficiency. Examples include the use of better optimizers \cite{rajbhandari2020zero,chen2023symbolic}, adopting distributed training strategies \cite{zhao2023pytorch}, employing mixed-precision training techniques \cite{micikevicius2017mixed}, and incorporating gradient accumulation \cite{chen2016training}. These efforts are complementary to our work.
\begin{figure*}[t!]
    \centering
    \begin{subfigure}[b]{0.66\columnwidth}
        \includegraphics[width=\columnwidth]{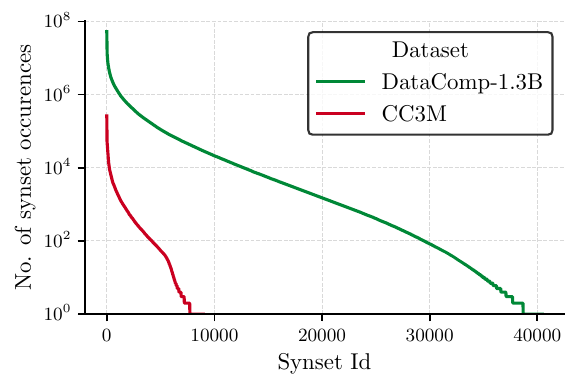}
        \caption{Synset distribution}
        \label{fig:synset_distribution}
    \end{subfigure}
    \hfill
    \begin{subfigure}[b]{0.66\columnwidth}
        \includegraphics[width=\columnwidth]{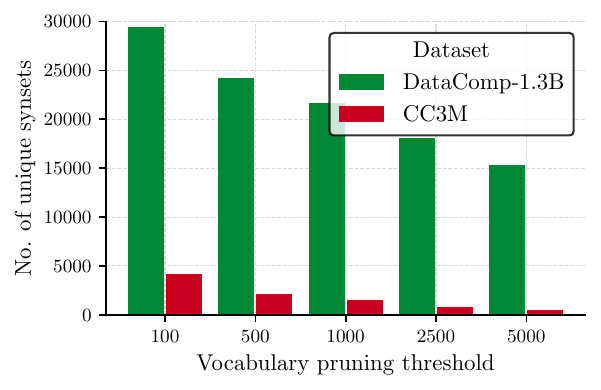}
        \caption{No. of synsets vs. $V_\tau$}
        \label{fig:syn_per_dataset}
    \end{subfigure}
    \hfill
    \begin{subfigure}[b]{0.66\columnwidth}
        \centering
        \resizebox{\columnwidth}{!}{
        \begin{tabular}{cc}
            \includegraphics[width=0.49\columnwidth]{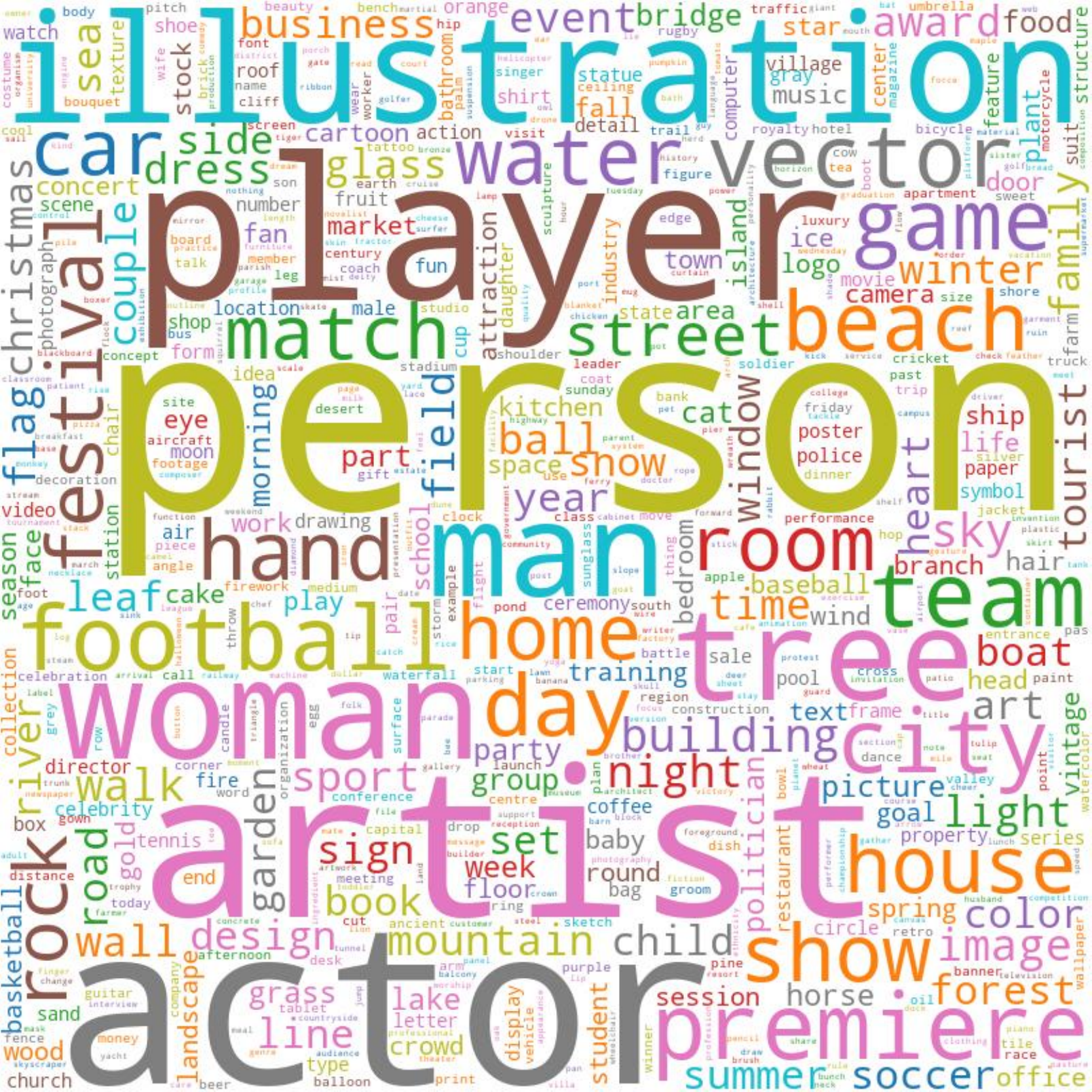} & \includegraphics[width=0.49\columnwidth]{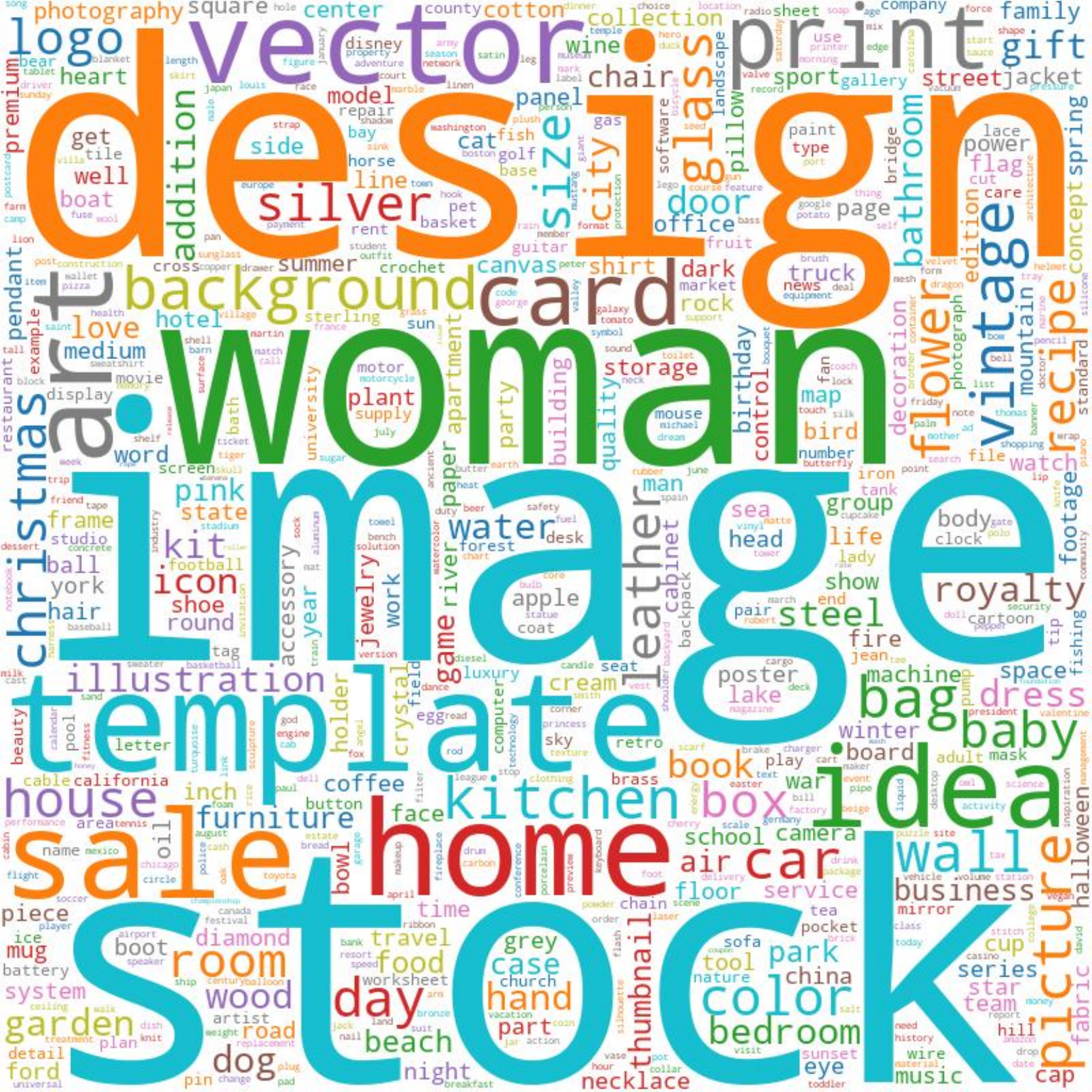}\\
            \cctm & \datacomp
            \end{tabular}
        }
        \caption{Top 1000 synset names}
        \label{}
    \end{subfigure}
    \caption{Analysis of extracted WordNet synsets in image-text datasets. Larger datasets typically contain a greater number of synsets, indicating increased content diversity in larger datasets.}
    \label{fig:data_analysis}
\end{figure*}

\section{\fontsize{11.5}{10}\selectfont{\method: Contrastive to Categorical Learning}}
\label{sec:simple}

For a given image-text dataset $\mathcal{D} = \left\{ (I^i, T^i)\ |\ i \in \left[1, N\right]\right\}$ with $N$ pairs of images $I$ and text captions $T$, CLIP aligns the image and text embeddings obtained from independent image and text encoders. This alignment is achieved through a contrastive loss which calculates pair-wise similarities between image and text embeddings. However, the computation of global pair-wise similarities is computationally demanding.

This paper casts image-text pre-training as a classification task. Our approach involves extracting nouns from text captions and mapping them to WordNet synsets \cite{miller1995wordnet} (\cref{ssec:caption_derived_labels}). We conduct pre-training on the \cctm dataset using a vision transformer model of \citet{dosovitskiy2020image}, and assess it's performance through linear probing accuracy on the \imagenet dataset. Unlike CLIP, the proposed method exhibits non-saturation with extended training (\cref{ssec:catlip_pretrianing}).

\subsection{Caption-derived classification labels}
\label{ssec:caption_derived_labels}

The abundance of noisy image-text data crawled from the web has enabled large-scale contrastive pre-training, leading to models improved generalization on visual recognition tasks. However, it is important to note that image-text pre-training is computationally intensive, particularly when compared with supervised pre-training for image classification on a similar scale of data (e.g., JFT-3B). On a positive note, the process of collecting image-text datasets at web-scale is more cost-effective compared to collecting labeled datasets. 

Leveraging the benefits of both web-scale image-text datasets and image classification frameworks effectively for faster training while preserving representation quality is an open question. We performed a simple analysis to answer this question. An overview of our approach is shown in \cref{fig:zero_shot_ann_app_main}. For a given text caption $T = \left(w_1, \cdots, w_m \right)$ with a sequence of $m$ words, we tag each word $w_i$ with part-of-speech tag ${pos}_i$, where $i$ is the position of word in a text caption. We then extract nouns and map them to WordNet synset as:
\begin{equation}
\resizebox{0.91\hsize}{!}{
    $
    \text{ExtractSynset}(T)= \{f(w_i)\ |\ pos_i \text{ is a noun}\, \forall i=1..m \}$
    }
    \label{eq:noun_synset}
\end{equation}
where $f$ is a function that maps $w_i$ to WordNet synset $S$.

To examine if the object labels of downstream classification dataset are present in image-text dataset $\mathcal{D}$, we extracted synset $\bar{S}$ using \cref{eq:noun_synset} for the object labels $y$ in the downstream classification dataset $\mathcal{\bar{D}} = \left\{ (\bar{I}^i, y^i)\ |\ i \in \left[1, M\right]\right\}$ with $M$ pairs of images $\bar{I}$ and object classification labels $y$. Synsets $S \in \mathcal{V}$ and $\bar{S}$ are similar if $sim(S, \bar{S}) \geq \alpha$, where $sim$ is a WordNet's path similarity  function and $\alpha$ is a pre-defined similarity threshold.

The statistics for different pre-training corpora are summarized in \cref{fig:data_analysis}. With the increase in pre-training image-text dataset size (from \cctm to \datacomp), the number of unique synsets increases (\cref{fig:syn_per_dataset}). This is expected, as a larger dataset typically contains a more diverse range of content, leading to a greater variety of unique objects and patterns. However, we observe that this also (1) increases the overlap with the labels in the downstream classification dataset (\ie, ImageNet-1k; \cref{fig:tau_500_main}) and (2) increases number of samples per synset (\cref{fig:in1k_syn_freq_pretraining_data_main}). This suggests that the larger pre-training dataset have more connections and commonalities with the labels in the downstream classification datasets, ultimately resulting in improvements in zero-shot accuracy. It's worth noting that because the model has been exposed to similar images of labels (or objects) in the downstream classification dataset during pre-training, \emph{manual template-based image retrieval (a.k.a., zero-shot image classification) in CLIP may not be truly zero-shot.} The notion of zero-shot image classification remains subject to debate. However, the analysis in \cref{fig:data_analysis} introduces an intriguing question: \emph{Can we cast image-text pre-training as a classification problem?}

\subsection{\method pre-training} 
\label{ssec:catlip_pretrianing}

We propose \method, a weakly supervised pre-training method based on WordNet synset supervision to answer this question. As shown in \cref{fig:overview_catlip}, \method extracts multiple synsets from input text captions and subsequently pretrains the image classification models using a binary cross-entropy loss, effectively shifting the problem from noisy image-text alignment to noisy image classification. 

\paragraph{Synset vocabulary.} The distribution of synsets in image-text datasets follows a long-tail distribution (\cref{fig:synset_distribution}). We create a vocabulary $\mathcal{V}: S \to C$ by counting the number of occurrences $C$ of the synset $S$ in given image-text dataset. Because some of the synsets will be less frequent, we prune $\mathcal{V}$ by storing synsets whose count $C$ is greater than a pre-defined vocabulary pruning threshold $V_\tau$. To determine a good value of $V_\tau$, we train \method with the base version of vision transformer (ViT B/16) as an image backbone on the \cctm dataset and evaluated the linear probe accuracy on the \imagenet dataset. Results in \cref{sec:app_additional_results} shows that $V_\tau = 100$ or $500$ is a good pruning threshold. In our experiments, we use $V_\tau=500$.

\begin{figure}[t!]
    \begin{subfigure}[t]{0.49\columnwidth}
        \centering
         \resizebox{\columnwidth}{!}{
        \includegraphics[width=\columnwidth]{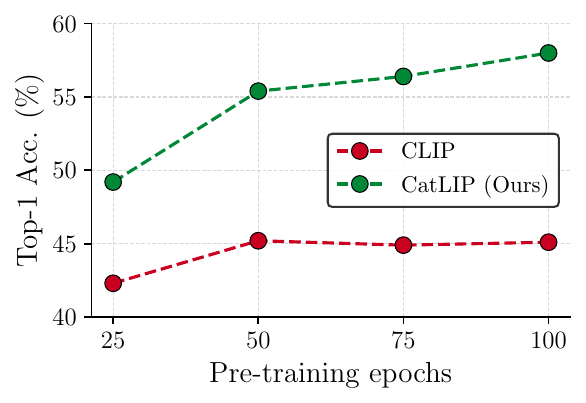}
        }
        \caption{Acc. vs. pre-training epochs}
        \label{fig:clip_vs_catlip_cc3m-epochs}
    \end{subfigure}
    \hfill
    %\vfill
    \begin{subfigure}[t]{0.49\columnwidth}
        \centering
         \resizebox{\columnwidth}{!}{
        \includegraphics[width=\columnwidth]{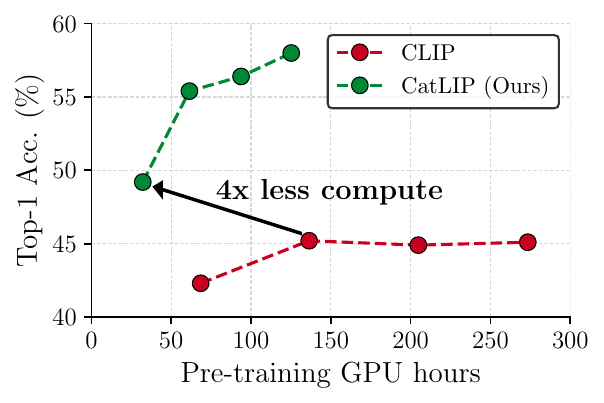}
        }
        \caption{Acc. vs. pre-training time}
        \label{fig:clip_vs_catlip_cc3m-time}
    \end{subfigure}    
    \caption{\textbf{\method vs. CLIP.} Unlike CLIP, \method benefits from longer training on CC3M dataset and requires significantly less pre-training computation. Here, (a) shows the top-1 linear probe accuracy of CLIP and \method on \imagenet as a function of CC3M pre-training epochs while (b) shows the linear probe accuracy as a function of pre-training duration. Here, each dot in a graph represents an independently trained model on CC3M. See \cref{sec:app_training_details} for training details.}
    \label{fig:clip_vs_catlip_cc3m}
\end{figure}

\paragraph{\method vs. CLIP.} Considering pre-training on image-text datasets as a classification problem, \emph{can \method serve as a viable alternative to CLIP pre-training?} To evaluate the comparative advantages of \method over CLIP, we conduct pre-training experiments with \method and CLIP on small-scale image-text dataset, i.e., \cctm, using ViT B/16 as an image backbone. We note that discussions on large-scale data and model scaling experiments are in \cref{sec:scale}.  We make following observations from results in \cref{fig:clip_vs_catlip_cc3m}.

First, the accuracy of the ViT B/16 model pre-trained with \method and linear probed on \imagenet improves as training is extended. In contrast, the accuracy of CLIP plateaus after a certain training duration. This observation aligns with previous works \citep[e.g.,][]{mu2022slip}, reinforcing the notion that CLIP requires large pre-training datasets to achieve optimal performance. 

Second, \method demonstrates faster pre-training as compared to CLIP. This is likely because \method optimizes only an image backbone during pre-training while CLIP optimizes both image and text backbones. Consequently, \method has several advantages over CLIP in terms of training efficiency: (1) it has faster step time (forward-backward pass) due to fewer network parameters to optimize, (2) requires less GPU memory for the same batch size, (3) benefits from GPU-optimized loss function implementations, and (4) has less communication overhead in multi-node training as it does not require synchronization between GPUs for computing global pair-wise similarities in contrastive loss. 

\section{Data and Model Scaling in \method}
\label{sec:scale}

Standard pre-training methods scale both data and model size to enhance accuracy. \cref{sec:simple} shows that \method is effective on small-scale datasets. This section studies the effect of data (\cref{ssec:data_scaling}) and model (\cref{ssec:model_scaling}) scaling on \method (see \cref{sec:app_training_details} for training details). We assess the quality of learned representations within the framework of transfer learning through (1) linear probing (LP), achieved by training a linear classifier on frozen image backbone weights and (2) full fine-tuning (FT), achieved by fine-tuning the entire model. We do LP and FT evaluations on two standard image classification datasets, i.e., \imagenet and \places \cite{zhou2017places}. For training details including hyper-parameters, see \cref{sec:app_training_details}.

\subsection{Data scaling} 
\label{ssec:data_scaling}
We pre-train \method with ViT B/16 as an image backbone on two publicly available image-text datasets (i.e., \cctm and \datacomp). \cref{tab:data_scaling} shows the linear probing (LP) and fine-tuning (FT) results of these pre-trained models. Increasing the data size from three million samples in \cctm to 1.3 billion samples in \datacomp improves the accuracy on transfer datasets. Consistent with previous works, this observation indicates that scaling data improves model's representation quality when pre-trained with \method. 

\begin{table}[t!]
    \centering
    \caption{\textbf{Data scaling in \method.} Scaling the image-text dataset from three million to 1.3 billion samples improves transfer learning accuracy of ViT B/16, both with linear probing (LP) and full fine-tuning (FT) on \imagenet and \places.}
    \label{tab:data_scaling}
    \vskip 0.15in
    \resizebox{\columnwidth}{!}{
    \begin{tabular}{lcccccc}
        \toprule[1.5pt]
        \multirow{2}{*}{\textbf{Dataset}} & \multirow{2}{*}{\textbf{\# samples}} & \multicolumn{2}{c}{\textbf{\imagenet Top-1 (\%)}} && \multicolumn{2}{c}{\textbf{\places Top-1 (\%)}} \\
        \cmidrule[1.25pt]{3-4}
        \cmidrule[1.25pt]{6-7}
        && \textbf{LP} & \textbf{FT} & & \textbf{LP} & \textbf{FT} \\
        \midrule[1.25pt]
        \cctm & 2.9 M & 58.0 & 67.3 && 49.1 & 54.5 \\
        \datacomp & 1.3 B & \textbf{80.1} & \textbf{84.3} && \textbf{53.6} & \textbf{59.2} \\
        \bottomrule[1.5pt]
    \end{tabular}
    }
    \vskip -0.1in
\end{table}

\begin{figure*}[t!]
    \centering
    \begin{subfigure}[b]{2.1\columnwidth}
        \centering
        \resizebox{\columnwidth}{!}{
        \begin{tabular}{ccc}
            ViT B/16 & ViT L/16 & ViT H/16 \\
             \includegraphics[width=0.33\columnwidth]{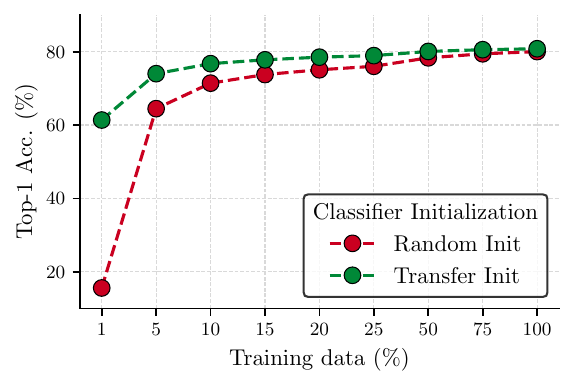} &  \includegraphics[width=0.33\columnwidth]{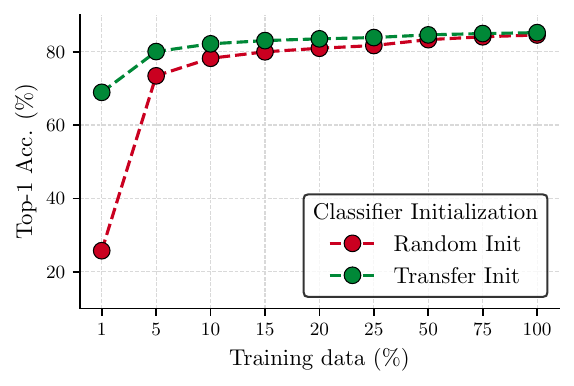} & \includegraphics[width=0.33\columnwidth]{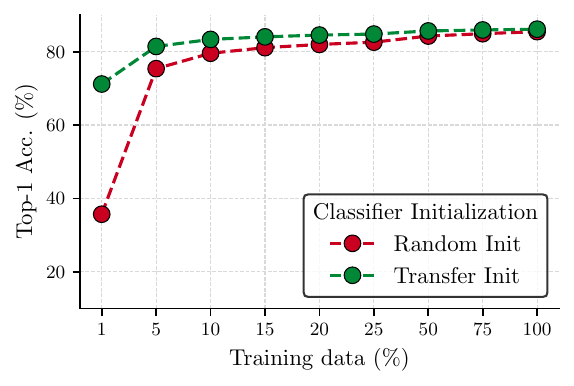}
        \end{tabular}
        }
        \caption{\imagenet}
        \label{fig:lp_results_in1k}
    \end{subfigure}
    \vfill
    \begin{subfigure}[b]{2.1\columnwidth}
        \centering
        \resizebox{\columnwidth}{!}{
            \begin{tabular}{ccc}
                 \includegraphics[width=0.33\columnwidth]{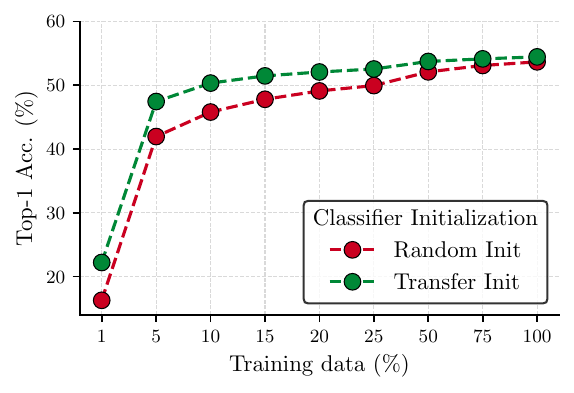} &  \includegraphics[width=0.33\columnwidth]{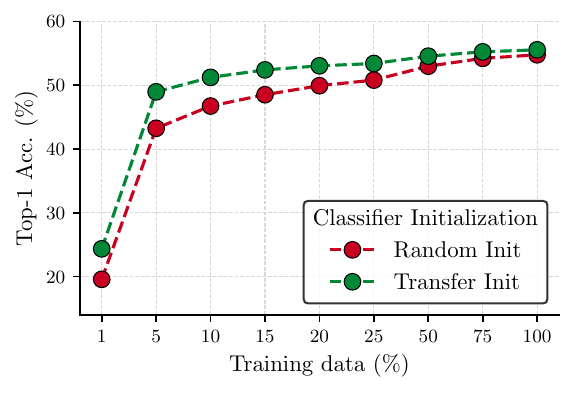} & \includegraphics[width=0.33\columnwidth]{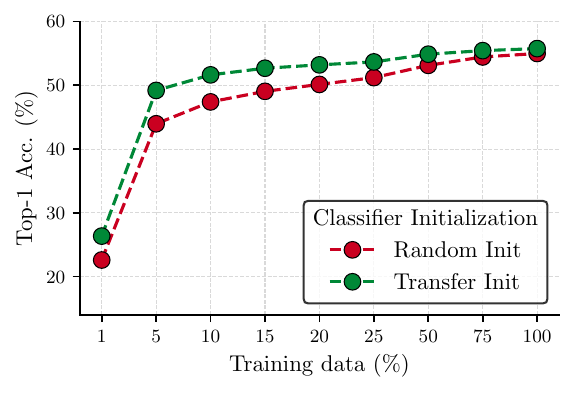}
            \end{tabular}
        }
        \caption{\places}
        \label{fig:lp_results_places}
    \end{subfigure}
    \caption{\textbf{Linear probe transfer learning with \method is more data efficient.} Here,  transfer learning is achieved by training a linear classifier on downstream tasks for 30 epochs with frozen image backbone features (each dot in a graph is an independent run). Linear classifier is initialized either using a standard method (Random Init) or transferred classifier embeddings from a pre-trained model (Transfer Init; ours). Initializing classifier with Transfer Init delivers better accuracy than Random Init, especially in small data regime.}
    \label{fig:data_efficiency_lp}
\end{figure*}

\begin{table*}[t!]
    \caption{Transfer learning accuracies of ViT models pre-trained on different datasets using supervised and weakly-supervised methods. Transfer learning is achieved through  fine-tuning the entire model on downstream classification tasks. Our weakly-supervised models achieve the best performance on both \imagenet and \places classification tasks. Here, we include models pre-trained on JFT under supervised pre-training as limited information is available about their semi-automatic labeling method, similar to \cite{singh2022revisiting}. $^\dagger$ ViT-22B uses frozen image encoder. Here, WIT means web-crawled image-text dataset. $^\star$ \gc{CoCa is a hybrid approach that uses labels from JFT-3B to create captions for image-text training along with ALIGN data. Therefore, it is not directly comparable to other approaches.}}
    \label{tab:compare_sota_pretraining}
    \vskip 0.15in
    \centering
    \resizebox{2\columnwidth}{!}{
    \begin{tabular}{lrcccccc}
        \toprule[1.5pt]
       \multirow{2}{*}{\textbf{Model}} & \multirow{2}{*}{\textbf{Params}} &  \multirow{2}{*}{\textbf{Pretraining}} & \multicolumn{2}{c}{\textbf{Resolution}}  && \multicolumn{2}{c}{\textbf{Top-1 accuracy}} \\
       \cmidrule[1.25pt]{4-5} \cmidrule[1.25pt]{7-8}
       & & & \textbf{Pre.} & \textbf{Fine.} && \textbf{\imagenet} & \textbf{\places} \\
       \midrule[1pt]
       \multicolumn{8}{l}{\emph{Supervised pre-training}} \\
       \midrule[1pt]
       ViT B/16 \cite{dosovitskiy2020image} & 87 M & ImageNet-21k & 224 & 384 && 84.0 & 58.2 \\
       ViT L/16 \cite{dosovitskiy2020image} & 305 M & ImageNet-21k & 224 & 384 && 85.2 & 59.0 \\
       ViT L/16 \cite{dosovitskiy2020image} & 305 M & JFT 300M & 224 & 512 && 87.8 &  \\
       ViT H/14 \cite{dosovitskiy2020image} & 634 M & JFT 300M & 224 & 512 && 88.6 & \\
       ViT L/16 \cite{zhai2022scaling} & 305 M & JFT 3B & 224 & 384 && 88.5 & \\
       ViT G/14 \cite{zhai2022scaling} & 1.9 B & JFT 3B & 224 & 518 && \textbf{90.5} & \\
       ViT-22B$^\dagger$ \cite{dehghani2023scaling} & 22 B & JFT 4B & 224 & 224 && 89.5 & \\
       \midrule[1pt]
       \multicolumn{8}{l}{\emph{Weakly supervised pre-training}} \\
       \midrule[1pt]
       ViT B/16 \cite{singh2022revisiting} & 87 M & IG 3.6B & 224 & 384 && 85.3 & 59.1 \\
       ViT L/16 \cite{singh2022revisiting} & 305 M & IG 3.6B & 224 & 512 && 88.1 & 60.7 \\
       ViT H/16 \cite{singh2022revisiting} & 633 M & IG 3.6B & 224 & 518 && 88.6 & 60.7 \\
       ALIGN EfficientNet-L2 \cite{jia2021scaling} & -- & ALIGN 1.8B & 289 & 600 && 88.6 &  \\
       FLIP ViT-B \cite{li2023scaling} & 329 M & LAION 2B & 224 & 224 && 87.1 & \\
       FLIP ViT-H \cite{li2023scaling} & 756 M & LAION 2B & 224 & 224 && 87.7 & \\
       OpenAI CLIP ViT B/16 \cite{radford2021learning} & 149 M & WIT 400M & 224 & 224 && 85.3 & \\
       RangeAugment CLIP ViT B/16 \cite{mehta2022rangeaugment} & 149 M & WIT 1.2B & 224 & 224 && 84.3 & \\
       RangeAugment CLIP ViT H/16 \cite{mehta2022rangeaugment} & 756 M & WIT 1.2B & 224 & 224 && 86.9 & \\
       OpenCLIP ViT B/16 \cite{cherti2023reproducible} & 149 M & LAION 2B & 224 & 224 && 85.5 & \\
       OpenCLIP ViT L/14 \cite{cherti2023reproducible} & 329 M & LAION 2B & 224 & 224 && 87.3 & \\
       OpenCLIP ViT H/14 \cite{cherti2023reproducible} & 756 M & LAION 2B & 224 & 224 && 87.6 & \\
       \gc{CoCa $^\star$ \cite{yu2022coca}} & \gc{2.1 B} & \gc{JFT 3B + ALIGN} & \gc{288} &  && \gc{91.0} & \\
       \hdashline
       CLIP ViT B/16 (Our repro.) & 149 M & \datacomp & 224 & 224 && 84.4 & 58.4 \\
       \method ViT B/16 (Ours) & 105 M & \datacomp & 224 & 224 && 84.3 & 59.2 \\
       \method ViT L/16 (Ours) & 329 M & \datacomp & 224 & 224 && 86.5 & 60.3 \\
       \method ViT H/16 (Ours) & 663 M & \datacomp & 224 & 224 && 86.7 & 60.2 \\
       \hdashline
       \method ViT B/16 (Ours) & 105 M & \datacomp & 224 & 512 && 86.1 & 59.4 \\
       \method ViT L/16 (Ours) & 329 M & \datacomp & 224 & 512 && 88.6 & 60.7 \\
       \method ViT H/16 (Ours) & 663 M & \datacomp & 224 & 512 && \textbf{88.8} & \textbf{61.1} \\
       \bottomrule[1.5pt]
    \end{tabular}
    }
    \vskip -0.1in
\end{table*}

\subsection{Model scaling}
\label{ssec:model_scaling}

We pre-train \method with different variants of vision transformer (ViT-B/16, ViT-L/16, and ViT-H/16) on the \datacomp dataset. \cref{fig:data_efficiency_lp} shows the linear probe evaluation on the \imagenet and \places datasets for different variants of ViT, respectively. We make following observations.

\paragraph{Representation quality improves with model size.} Scaling up the model size improves the representation quality. However, the performance of the largest model (i.e., ViT H/16) starts to saturate on \method (ViT B/16 vs. ViT L/16 vs. ViT H/16: 80.8 vs. 85.2 vs. 86.1; \cref{fig:lp_results_in1k}). These results on \imagenet are consistent with scaling laws for ViT on large-scale supervised datasets \cite{zhai2022scaling}. We also observe similar trends on \places dataset (\cref{fig:lp_results_places}).

\paragraph{Better transfer learning with bigger models.} Linear probing on task-specific datasets shows that bigger models exhibit better transfer learning capabilities, especially in small data regime (red color curves in \cref{fig:data_efficiency_lp}). For example, the linear probe top-1 accuracy of ViT B/16, ViT L/16, and ViT H/16 when trained with 1\% of \imagenet data is about 16\%, 26\%, and 36\% respectively. This highlights their capacity to leverage pre-existing knowledge and generalize effectively on small datasets.

\paragraph{Data efficient transfer learning.} Many target labels in downstream classification datasets are English nouns. Therefore, we can map target labels and the synset vocabulary $\mathcal{V}$ based on their WordNet path similarity score, as shown in \cref{fig:zero_shot_ann_app_main}.  Subsequently, extracting the embeddings associated with target labels from the classification layer of the pre-trained model, for initializing the classification layer of the target task, is a natural and logical step. We refer to this classifier initialization as \emph{Transfer Init} and compare it with random classifier initialization (\emph{Random Init}).

We make two observations from results in \cref{fig:data_efficiency_lp}: (1) In the context of the \imagenet dataset, all downstream task labels are subset of $\mathcal{V}$ (\cref{fig:tau_500_main}). Consequently, Transfer Init yields significantly better accuracy, especially in small data regimes (\cref{fig:lp_results_in1k}). This performance improvement aligns with expectations as the pre-training on large corpora has exposed the model to images of similar objects. (2) In the case of the \places dataset, Transfer Init continues to result in improved accuracy, however, gains are less substantial compared to observation (1).
This disparity is attributed to the fact that \places have target labels that are either not a subset of $\mathcal{V}$ (e.g., \texttt{archaeological\_excavation})\footnote{Initialized with random embeddings.} or comprise a combination of multiple synsets in $\mathcal{V}$ (e.g., \texttt{restaurant\_kitchen})\footnote{Initialized with an average embedding.}. As a consequence,  model requires more data to adapt to these unseen labels or linearly interpolated embeddings, resulting in a less pronounced improvement compared to more aligned labels in ImageNet.

\subsection{Comparison with existing pre-training methods}
Our weakly supervised method, \method, is compared with state-of-the-art methods in \cref{tab:compare_sota_pretraining}. We compare pre-training parameter count, pre-training data size, image resolution during pre-training and fine-tuning, and top-1 accuracy on two standard datasets: \imagenet and \places. We group these methods into two categories: supervised and weakly-supervised. We classify models pre-trained on JFT as supervised because limited information is available about these datasets and their annotation process. \cite{zhai2022scaling} briefly mentions that JFT-3B has 30k classes, and is collected using a semi-automatic annotation pipeline, implying a manual curation and annotation process. Therefore, we consider JFTs as a supervised dataset, similar to \cite{singh2022revisiting}.

\cref{tab:compare_sota_pretraining} shows that \method delivers competitive performance compared to existing pre-training methods. For instance, ViT B/16 pre-trained with \method and CLIP on \datacomp achieves 84.3\% and 84.4\% top-1 accuracy on \imagenet, and 58.4\% and 59.2\% on \places respectively. It is important to note that pre-training with \method is $2.7\times$ faster than CLIP (\cref{fig:catlip_clip_compare_main_results}). These results suggest that \method is an efficient and effective alternative to contrastive pre-training on large-scale image-text datasets. 

Additionally, ViT variants trained with \method are competitive to other publicly available CLIP variants, such as OpenCLIP \cite{cherti2023reproducible} and RangeAugment CLIP \cite{mehta2022rangeaugment}, despite being pre-trained on different datasets. Also, comparison with \citep{singh2022revisiting} is interesting because it is also trained as a multi-label classifier on Instagram images, utilizing hashtags as labels for the images. Remarkably, \method achieves similar  performance to \citep{singh2022revisiting}, but with about $2.8\times$ smaller pre-training data. 

\section{Task Generalization}
\label{sec:generalize}
Many vision pre-training methods commonly use a single-label classification benchmark (e.g., \imagenet) for evaluation. However, real-world images are inherently context-rich. These images often comprises of complex backgrounds, intricate scenes, and multiple objects, presenting a more nuanced and detailed visual environment than what is typically addressed in single-label classification tasks (e.g., \imagenet). 

To assess the effectiveness of \method in  such real-world scenarios, we conduct experiments on three challenging tasks: (1) multi-label object classification (\cref{ssec:multi_label_obj}), (2) semantic segmentation (\cref{ssec:semantic_seg}), and (3) object detection (\cref{ssec:object_detection}). Our observations show that models pre-trained with \method achieve similar or better accuracy compared to CLIP on these tasks. This emphasizes that \method effectively learns high-quality representations, showcasing its versatility and robustness in addressing complex visual recognition scenarios.

\subsection{Multi-label object classification} 
\label{ssec:multi_label_obj}

We evaluate the performance of models pre-trained with \method on the multi-object classification task using the \coco dataset. This dataset consists of approximately 118,000 training images categorized into 80 classes, with an average of around 3 labels per image. We evaluate the performance in terms of mean average precision (mAP) on the validation set, consisting of 5,000 images.

Table \ref{tab:coco_cls_results} shows that \method achieves comparable performance to CLIP. Remarkably, \method, trained with a simple binary cross-entropy loss, delivers competitive accuracy compared to existing methods employing more complex techniques, such as asymmetric loss functions \cite{ridnik2023ml} and pyramidal feature extraction \cite{xu2023open}.

\begin{table}[t!]
    \caption{\textbf{Multi-label object classification on \coco.} For reference, we've included results from state-of-the-art methods with complex architectures and loss functions in gray cells. Here, $^\dagger$ denotes that results are from \cite{xu2023open}.}
    \label{tab:coco_cls_results}
    \vskip 0.15in
    \centering
    \resizebox{0.85\columnwidth}{!}{
    \begin{tabular}{lcc}
        \toprule[1.5pt]
       \textbf{Model} & \textbf{Image Encoder} & \textbf{mAP (in \%)}  \\
        \midrule[1.25pt]
         \gc{ML-Decoder$^\dagger$} \cite{ridnik2023ml} & \gc{ViT L} & \gc{90.6} \\
         \gc{ADDS} \cite{xu2023open} & \gc{ViT L} & \gc{91.8} \\
         \midrule
        CLIP (Our impl.) & ViT B/16 & 87.4 \\
        \hdashline
        \multirow{2}{*}{\method (Ours)} & ViT B/16 & 87.9 \\
        & ViT L/16 & 90.6 \\
         \bottomrule[1.5pt]
    \end{tabular}
    }
    \vskip -0.1in
\end{table}

\subsection{Semantic segmentation}
\label{ssec:semantic_seg}

We evaluate the semantic segmentation performance on ADE20k \cite{zhou2017scene}. ADE20k comprises 150 semantic categories with approximately 20,000 training and 2,000 validation images. We use Deeplabv3 \cite{chen2017rethinking} as a segmentation head. We use mean intersection over union (mIoU) metric to measure accuracy.

%In contrast, COCO-Stuff includes 172 categories (80 things, 91 stuff, and 1 unlabeled class) and has around 118,000 training and 5,000 validation images. 

The results in \cref{tab:semantic_seg} show that DeepLabv3 with \method's ViT B/16 as an image encoder achieves competitive performance compared to CLIP's ViT B/16. Interestingly, DeepLabv3 with \method's ViT image encoders delivers similar or better performance than existing methods that employ more complex segmentation heads. For example, \method's ViT L/16 with DeepLabv3 as segmentation head is about 1\% more accurate than ViT Adaptor \cite{chen2022vision} with UpperNet \cite{xiao2018unified} segmentation head with similar number of network parameters.

\begin{table}[t]
    \caption{\textbf{Semantic segmentation on ADE20k.} For reference, we have included results from state-of-the-art methods (MMSeg \cite{mmseg2020}, SegViT \cite{zhang2022segvit}, and ViT Adaptor \cite{chen2022vision}) with complex architectures (UpperNet \cite{xiao2018unified} and Semantic FPN \cite{kirillov2019panoptic}) in gray.}
    \label{tab:semantic_seg}
    \vskip 0.15in
    \centering
    \resizebox{\columnwidth}{!}{
    \begin{tabular}{lccrccccc}
        \toprule[1.5pt]
        \textbf{Model} & \textbf{Image encoder} & \textbf{Seg. head} & \textbf{\# params.} & \textbf{Resolution} &\textbf{mIoU} \\
        \midrule[1pt]
        \gc{MMSeg} & \gc{ViT B} & \gc{UpperNet} & \gc{144 M} & \gc{$512 \times 512$} & \gc{47.73} \\
        \gc{SegViT} & \gc{ViT L} & \gc{SegViT} & \gc{344 M} & \gc{$640 \times 640$} & \gc{54.60} \\
        \gc{ViT Adaptor} & \gc{ViT L} & \gc{Semantic FPN} & \gc{332 M} & \gc{$512 \times 512$} & \gc{52.90} \\
        \gc{ViT Adaptor} & \gc{ViT L} & \gc{UpperNet} & \gc{332 M} & \gc{$512 \times 512$} & \gc{53.40} \\
        \midrule
        CLIP (Our impl.) & ViT B/16 & DeepLabv3 & 99 M & $512 \times 512$ & 49.70 \\
        \hdashline
        \multirow{3}{*}{\method} & ViT B/16 & \multirow{3}{*}{DeepLabv3} & 99 M & \multirow{3}{*}{$512 \times 512$}  & 50.10 \\
                & ViT L/16 & & 320 M && 54.46 \\
                & ViT H/16 & & 652 M &&  \textbf{55.63} \\
        \bottomrule[1.5pt]
    \end{tabular}
    }
    \vskip -0.1in
\end{table}

\subsection{Object detection and instance segmentation}
\label{ssec:object_detection}
We evaluate the object detection and instance segmentation performance on the \coco dataset with Mask R-CNN as detection head. The dataset comprises of 118,000 training and 5000 validation images. We measure the accuracy in terms of mean average precision metric (mAP @ IoU of 0.50:0.95).

Results are given in \cref{tab:object_detection}. We see that Mask R-CNN with \method's ViT B/16 image backbone delivers similar performance to CLIP's ViT B/16 backbone. We also observe that Mask RCNN's mAP increases when we scale image backbone from ViT B/16 to ViT H/16, and we obtain results similar to existing method \cite{li2022exploring}. These results show that \method learns high quality representations. 

\begin{table}[t!]
    \caption{\textbf{Object detection and instance segmentation on \coco with Mask R-CNN.} As a reference and for fair comparison, we include results from ViTDet \cite{li2022exploring} without any information propagation strategies.}
    \label{tab:object_detection}
    \vskip 0.15in
    \centering
    \resizebox{0.75\columnwidth}{!}{
    \begin{tabular}{lcccc}
        \toprule[1.5pt]
        \multirow{2}{*}{\textbf{Model}} & \textbf{Image} && \multicolumn{2}{c}{\textbf{\coco}} \\
        \cmidrule[1.25pt]{4-5}
        & \textbf{encoder} && \textbf{AP\textsuperscript{Box}} & \textbf{AP\textsuperscript{Mask}} \\
        \midrule[1pt]
        \multirow{2}{*}{\gc{ViTDet}} & \gc{ViT B} && \gc{48.9} & \gc{43.9} \\
             & \gc{ViT L} && \gc{52.9} & \gc{47.2} \\
        \midrule
        CLIP (Our impl.) & ViT B/16 && 49.9 & 43.7 \\
        \hdashline
        \multirow{3}{*}{\method} & ViT B/16 && 49.9 & 43.6 \\
                & ViT L/16 && 52.7 & 46.0 \\
                & ViT H/16 && 53.4 & 46.4 \\
        \bottomrule[1.5pt]
    \end{tabular}
    }
    \vskip -0.1in
\end{table}
\section{Conclusion}
\label{sec:conclusion}

This paper introduces a weakly-supervised pre-training method for vision models using publicly available web-scale image-text data. By recasting pre-training as a classification task, the proposed method addresses the computational challenges associated with contrastive learning, leading to an impressive $2.7\times$ training speed-up on web-scale data while preserving transfer learning accuracies across varieties of visual recognition tasks, including detection and segmentation. This work contributes significantly to the efficient and effective pre-training of vision models. We hope this work will facilitate efficient pre-training research on web-scale noisy data.
\section*{Broader Impact}

Large-scale pre-training on image-text datasets have led to a series of breakthroughs in computer vision and related fields. However, large-scale pre-training is still challenging because it requires significant computational resources. This paper provides insights into the efficiency of image-text pre-training, and proposes a reformulation with significantly lower computational requirements. We show the training efficiency gains while preserving transfer accuracies on several visual recognition tasks. We will open-source our code and pre-trained models to enable future research. We hope that our work will make research in pre-training accessible. Otherwise, we expect that it will have the same broader impact, both positive and negative, as the field as a whole.

\bibliography{main}
\bibliographystyle{icml2024}

\newpage
\appendix
\onecolumn

\section{Training Details}
\label{sec:app_training_details}

\paragraph{Pre-training on image-text datasets.} \cref{tab:pre_training_hyerparams} lists the hyper-parameters used for pre-training on \cctm and \datacomp with \method. Following \cite{mehta2022rangeaugment}, we use a multi-scale variable batch sampler \cite{mehta2021mobilevit} with a base image resolution of $224 \times 224$. We use AdamW \cite{loshchilov2017decoupled} with default $\beta$ values as an optimizer and binary cross-entropy loss as an objective function. We use cosine learning rate schedule \cite{loshchilov2016sgdr}. For augmentation, we use random resized crop with default scale and aspect ratios, random horizontal flipping, and RangeAugment \cite{mehta2022rangeaugment}. 

\begin{table}[b!]
    \centering
    \caption{Pre-training hyper-parameters for CLIP and \method.}
    \label{tab:pre_training_hyerparams}
    %\vskip 0.15in
    \begin{subtable}[t]{0.4\columnwidth}
        \centering
        \caption{\cctm experiments in \cref{fig:clip_vs_catlip_cc3m}}
        \label{tab:app_cc3m_hp}
        \resizebox{!}{60px}{
        \begin{tabular}{lcc}
            \toprule[1.5pt]
            \textbf{Hyperparameter} & \textbf{CLIP} & \textbf{\method} \\
            \midrule[1pt]
            Backbone & ViT B/16 & ViT B/16 \\
            Max. Epochs & 25/50/75/100 & 25/50/75/100 \\
            LR scheduler & Cosine & Cosine \\
            Max. LR & 0.0005 & 0.002 \\
            Min. LR & 0.000001 & 0.00002 \\
            Optimizer & AdamW & AdamW \\
            AdamW $\beta$'s & (0.9, 0.98) & (0.9, 0.999) \\
            Weight decay & 0.2 & 0.2 \\
            Batch size per GPU & 512 & 512 \\
            \# A100 GPUs & 4 & 4 \\
            A100 GPU Memory & 40 GB & 40 GB \\
            \bottomrule[1.5pt]
        \end{tabular}
        }
    \end{subtable}
    \hfill
    \begin{subtable}[t]{0.58\columnwidth}
        \centering
        \caption{\datacomp}
        \label{tab:app_datacomp_hp}
        \resizebox{!}{60px}{
        \begin{tabular}{lcccc}
            \toprule[1.5pt]
            \textbf{Hyperparameter} & \textbf{CLIP} & \multicolumn{3}{c}{\textbf{\method}} \\
            \cmidrule[1.25pt]{3-5}
             & ViT B/16 & ViT B/16 & ViT L/16 & ViT H/16 \\
            \midrule[1pt]
            Max. Iterations & 200000 & 200000 & 200000 & 200000 \\
            LR scheduler & Cosine & Cosine & Cosine & Cosine\\
            Max. LR & 0.0005 & 0.001 & 0.0006 & 0.0004 \\
            Min. LR & 0.000001 & 0.00001 & 0.000006 & 0.000004\\
            Optimizer & AdamW & AdamW & AdamW & AdamW \\
            AdamW $\beta$'s & (0.9, 0.98) & (0.9, 0.999) & (0.9, 0.999) & (0.9, 0.999) \\
            Weight decay & 0.2 & 0.2 & 0.2 & 0.2 \\
            Batch size per GPU & 256 & 1024 & 512 & 256 \\
            \# A100 GPUs & 256 & 64 & 128 & 256 \\
            A100 GPU Memory & 80 GB & 80 GB & 40 GB & 40 GB \\
            \bottomrule[1.5pt]
        \end{tabular}
        }
    \end{subtable}
    \vskip -0.1in
\end{table}

\begin{table}[b!]
    \centering
    \caption{Transfer learning hyper-parameters image classification tasks. Here, LP and FT means linear probing and full finetuning respectively.}
    \label{tab:transfer_learning_single_label_cls}
    %\vskip 0.15in
    \begin{subtable}[t]{0.32\columnwidth}
        \centering
        \caption{LP on \imagenet and \places.}
        \label{tab:app_lp_in_places}
        \resizebox{0.8\columnwidth}{!}{
        \begin{tabular}{lc}
            \toprule[1.5pt]
            \textbf{Hyperparameter} & \textbf{Value} \\
            \midrule[1pt]
            Backbone & ViT B/L/H \\
            Epochs & 30 \\
            LR scheduler & Cosine \\
            Max. LR & 0.00003 \\
            Min. LR & 0.000001 \\
            Optimizer & AdamW \\
            Weight decay & 0 \\
            Batch size per GPU & 256 \\
            \# A100 GPUs & 1 \\
            A100 GPU Memory & 80 GB \\
            \bottomrule[1.5pt]
        \end{tabular}
        }
    \end{subtable}
    \hfill
    \begin{subtable}[t]{0.32\columnwidth}
        \centering
        \caption{FT on \imagenet and \places.}
        \label{tab:app_ft_in_places}
        \resizebox{0.8\columnwidth}{!}{
        \begin{tabular}{lc}
            \toprule[1.5pt]
            \textbf{Hyperparameter} & \textbf{Value} \\
            \midrule[1pt]
            Backbone & ViT B/L/H \\
            Epochs & 10 \\
            LR scheduler & Cosine \\
            Max. LR & 0.00003 \\
            Min. LR & 0.000003 \\
            Optimizer & AdamW \\
            Weight decay & 0.05 \\
            Batch size per GPU & 128 \\
            \# A100 GPUs & 1 \\
            A100 GPU Memory & 80 GB \\
            \bottomrule[1.5pt]
        \end{tabular}
        }
    \end{subtable}
    \hfill
    \begin{subtable}[t]{0.32\columnwidth}
        \centering
        \caption{FT on \coco.}
        \label{tab:app_ft_coco_cls}
        \resizebox{0.8\columnwidth}{!}{
        \begin{tabular}{lc}
            \toprule[1.5pt]
            \textbf{Hyperparameter} & \textbf{Value} \\
            \midrule[1pt]
            Backbone & ViT B/L/H \\
            Epochs & 10 \\
            LR scheduler & Cosine \\
            Max. LR & 0.00001 \\
            Min. LR & 0.000001 \\
            Optimizer & AdamW \\
            Weight decay & 0.05 \\
            Batch size per GPU & 32 \\
            \# A100 GPUs & 1 \\
            A100 GPU Memory & 80 GB \\
            \bottomrule[1.5pt]
        \end{tabular}
        }
    \end{subtable}
    \vskip -0.1in
\end{table}

\paragraph{Transfer learning on image classification.} \cref{tab:transfer_learning_single_label_cls} lists the hyper-parameters used for pre-training on \imagenet, \places, and \coco. We use a multi-scale variable batch sampler with a base image resolution of $224 \times 224$. We use AdamW  with default $\beta$ values as an optimizer. For single-label classification (\imagenet and \places), we use cross-entropy loss with label smoothing as an objective function. For multi-label classification (\coco), we use binary cross entropy as an objective function. We use cosine learning rate schedule. For high resolution experiments in \cref{tab:compare_sota_pretraining} and multi-label classification experiments in \cref{tab:coco_cls_results}, we use $512 \times 512$ as the base image resolution.

\paragraph{Transfer learning on semantic segmentation and object detection.} \cref{tab:transfer_learning_single_label_cls} lists the hyper-parameters used for pre-training on ADE20k for semantic segmentation and \coco for object detection. For segmentation, we use random short size resize, random crop, and horizontal flipping as augmentation along with RangeAugment. We use a fixed-size batch sampler with a base image resolution of $512 \times 512$. We use AdamW  with default $\beta$ values as an optimizer. We use cross-entropy as an objective function and cosine schedule for annealing learning rate. 

For detection, we use the same augmentations as used in ViTDet \cite{li2022exploring}.  We use a multi-scale variable batch sampler with a base image resolution of $1024 \times 1024$. We use AdamW  with default $\beta$ values as an optimizer. We use the same loss function as used in ViTDet for regressing bounding box regression and object name classifcation. For learning rate annealing, we use multi-step learning rate schedule.

\begin{table}[t!]
    \centering
    \caption{Transfer learning hyper-parameters for semantic segmentation and object detection.}
    \label{tab:transfer_learning_seg_det}
    \begin{subtable}[t]{0.42\columnwidth}
        \centering
        \caption{Semantic segmentation on ADE20k.}
        \label{tab:app_seg_ade20k_hp}
        \resizebox{0.8\columnwidth}{!}{
        \begin{tabular}{lc}
            \toprule[1.5pt]
            \textbf{Hyperparameter} & \textbf{Value} \\
            \midrule[1pt]
            Seg. Head & DeepLabv3 \\
            Image Backbone & ViT B/L/H \\
            Epochs & 50 \\
            LR scheduler & Cosine \\
            Max. LR & 0.00003 \\
            Min. LR & 0.000003 \\
            Optimizer & AdamW \\
            Weight decay & 0.1 \\
            Batch size per GPU & 4 \\
            \# A100 GPUs & 8 \\
            A100 GPU Memory & 80 GB \\
            \bottomrule[1.5pt]
        \end{tabular}
        }
    \end{subtable}
    \hfill
    \begin{subtable}[t]{0.55\columnwidth}
        \centering
        \caption{Object detection on \coco.}
        \label{tab:app_mask_rcnn_hp}
        \resizebox{0.8\columnwidth}{!}{
        \begin{tabular}{lc}
            \toprule[1.5pt]
            \textbf{Hyperparameter} & \textbf{Value} \\
            \midrule[1pt]
            Det. Head & Mask R-CNN \\
            Image Backbone & ViT B/L/H \\
            Epochs & 25/25/20 \\
            Optimizer & AdamW \\
            Weight decay & 0.1 \\
            LR scheduler & Multi-step \\
            Max. LR & 0003 \\
            LR step size & 0.1 \\
            LR step interval & (22, 24) / (22, 24) / (17, 19) \\
            Layer-wise LR decay & 0.7/0.8/0.9 \\
            Stochastic dropout & 0.1/0.4/0.5 \\
            Batch size per GPU & 2 \\
            \# A100 GPUs & 64 \\
            A100 GPU Memory & 80 GB \\
            \bottomrule[1.5pt]
        \end{tabular}
        }
    \end{subtable}
    \vskip -0.1in
\end{table}

\section{Additional results}
\label{sec:app_additional_results} 

\begin{figure}[b!]
    \centering
    \includegraphics[width=0.4\columnwidth]{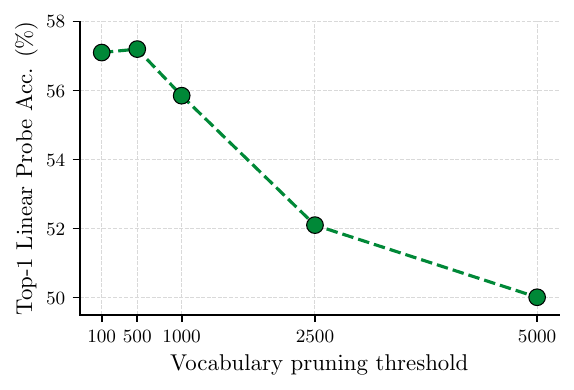}
    \caption{Linear probing accuracy on \imagenet as a function of vocabulary pruning threshold $V_\tau$. Here, each data point represents a ViT B/16 model pretrained on \cctm at different values of vocabulary pruning thresholds $V_\tau$.}
    \label{fig:app_vit_b_top1_vs_threshold_cc3m}
\end{figure}

\paragraph{Effect of vocabulary pruning threshold.} The distribution of synsets in image-text datasets demonstrates a long-tail pattern, necessitating the pruning of infrequently occurring synsets. In \cref{fig:app_vit_b_top1_vs_threshold_cc3m}, we show the results obtained by pruning the synset vocabulary $\mathcal{V}$ at various thresholds, denoted as $V_\tau$. For each value of $V_\tau$, we train an independent model using \method on \cctm. Here, we use ViT-B as the image encoder. 

To determine the optimal pruning threshold, we conduct linear probing experiments on \imagenet. Notably, larger values of $V_\tau$ lead to a significant decline in performance, likely attributed to the removal of crucial information about objects at such elevated thresholds. We found that $V_\tau=100$ or $V_\tau=500$ are good pruning thresholds. Therefore, in our experiments involving both \cctm and \datacomp, we use $V_\tau=500$ as the threshold.

\begin{figure}[t!]
    \centering
        \begin{subfigure}[b]{0.33\columnwidth}
            \includegraphics[width=\columnwidth]{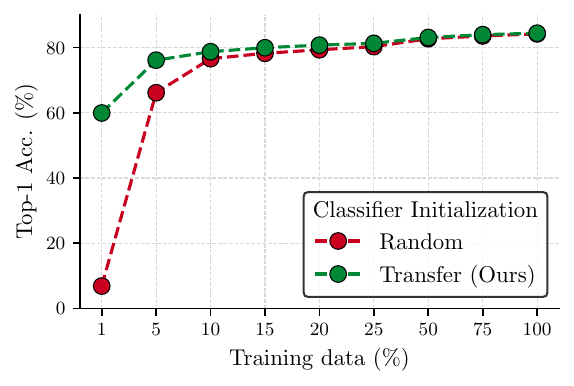}
            \caption{ViT-B}
        \end{subfigure}
        \begin{subfigure}[b]{0.33\columnwidth}
            \includegraphics[width=\columnwidth]{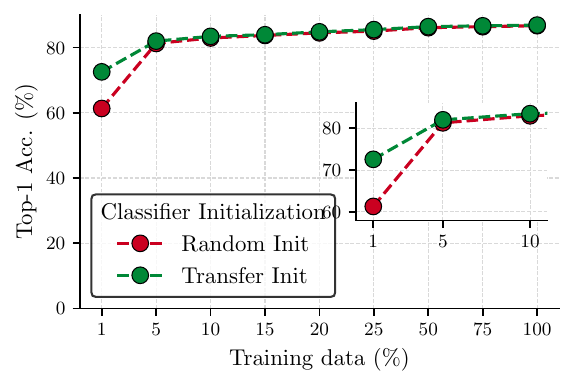}
            \caption{ViT-L}
        \end{subfigure}
        \begin{subfigure}[b]{0.33\columnwidth}
            \includegraphics[width=\columnwidth]{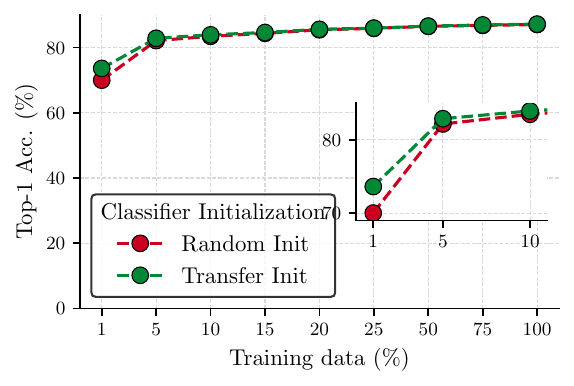}
            \caption{ViT-H}
        \end{subfigure}
    \caption{Transfer accuracy of models pre-training using \method on \imagenet. Transfer learning is achieved by finetuning the entire model on \imagenet for 10 epochs. Linear classifier is initialized either using a standard method (Random Init) or transferred classifier embeddings from a pre-trained model (Transfer Init). Transfer Init delivers better accuracy than Random Init, especially in small data regimes.}
    \label{fig:app_ft_in1k}
\end{figure}

\paragraph{Data efficient transfer learning.} \cref{sec:scale} highlights the effectiveness of Transfer Init, showcasing its effictiveness in transferring information from the pre-trained classification layer using \method to the downstream task's classification layer through linear probing on \imagenet. In \cref{fig:app_ft_in1k}, a comparable trend to linear probing is observed when finetuning the entire model on \imagenet at varying data percentages. However, these trends are not as pronounced as in linear probing, as anticipated. This discrepancy can be attributed to the backbone's adaptability to new labeled data during fine-tuning, in contrast to the frozen backbone in linear probing.

\begin{figure}[t!]
    \centering
    \includegraphics[width=0.4\columnwidth]{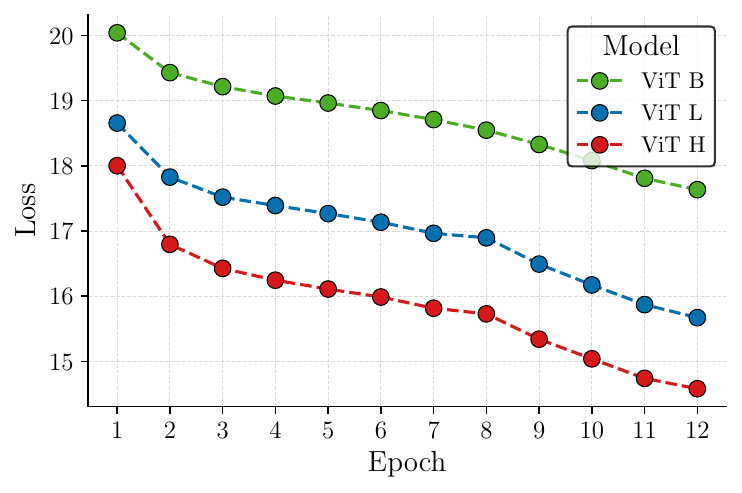}
    \caption{Training loss vs. epoch for different variants of ViT models, pre-trained with \method on \datacomp.}
    \label{fig:app_pretraining_loss}
\end{figure}

\paragraph{Pre-training loss.} Training loss during pre-training on \datacomp is shown in \cref{fig:app_pretraining_loss} for different variants of ViT. Loss decreases with increase in model size. This observed correlation aligns with the transfer learning accuracy of different ViT models on downstream tasks (\cref{sec:scale,sec:generalize}). Importantly, the loss is not saturated, suggesting that longer training will further improve performance.

\end{document}